\documentclass[10pt,twocolumn,letterpaper]{article}

\usepackage{cvpr}              

\usepackage{float} 
\usepackage[normalem]{ulem}
\usepackage{tikz}
\usepackage{multirow}
\usepackage{float} 
\usepackage[table]{xcolor} 

\definecolor{cvprblue}{rgb}{0.21,0.49,0.74}
\usepackage[pagebackref,breaklinks,colorlinks,allcolors=cvprblue]{hyperref}

\usepackage{amsmath}
\usepackage{graphicx}
\def\paperID{15847} 
\def\confName{CVPR}
\def\confYear{2026}

\title{Uni-Hema: Unified Model for Digital Hematopathology }

\author{
Abdul Rehman${}^{1}$,
Iqra Rasool${}^{2}$,
Ayesha Imran${}^{2}$,
Mohsen Ali${}^{1}$,
Waqas Sultani${}^{1}$ \\
\\
${}^{1}$Information Technology University of Punjab, Lahore \\
${}^{2}$Chughtai Lab, Lahore \\
{\tt\small phdcs23002@itu.edu.pk, mohsen.ali@itu.edu.pk, waqas.sultani@itu.edu.pk}
}

\begin{document}
\maketitle
\begin{abstract}
Digital hematopathology requires cell-level analysis across diverse disease categories, including malignant disorders (e.g., leukemia), infectious conditions (e.g., malaria), and non-malignant red blood cell disorders (e.g., sickle cell disease). Whether single-task, vision-language, WSI-optimized, or single-cell hematology models, these approaches share a key limitation: they cannot provide unified, multi-task, multi-modal reasoning across the complexities of digital hematopathology. To overcome these limitations, we propose \textbf{Uni-Hema}, a multi-task, unified model for digital hematopathology integrating detection, classification, segmentation, morphology prediction, and reasoning across multiple diseases. 
Uni-Hema leverages 46 publicly available datasets, encompassing over 700K images and 21K question-answer pairs, and is built upon \textbf{Hema-Former}, a multimodal module that bridges visual and linguistic representations at the hierarchy level for the different tasks (detection, classification, segmentation, morphology, mask language modeling and visual question answer) at different granularity. Extensive experiments demonstrate that Uni-Hema achieves comparable or superior performance to train on a single-task and single dataset models,  across diverse hematological tasks, while providing interpretable, morphologically relevant insights at the single-cell level. 
Our framework establishes a new standard for multi-task and multi-modal digital hematopathology. 
The code will be made publicly available.

\end{abstract}

\section{Introduction}
\label{sec:intro}
\begin{figure}[h]
    \centering
\includegraphics[width=0.9\columnwidth]{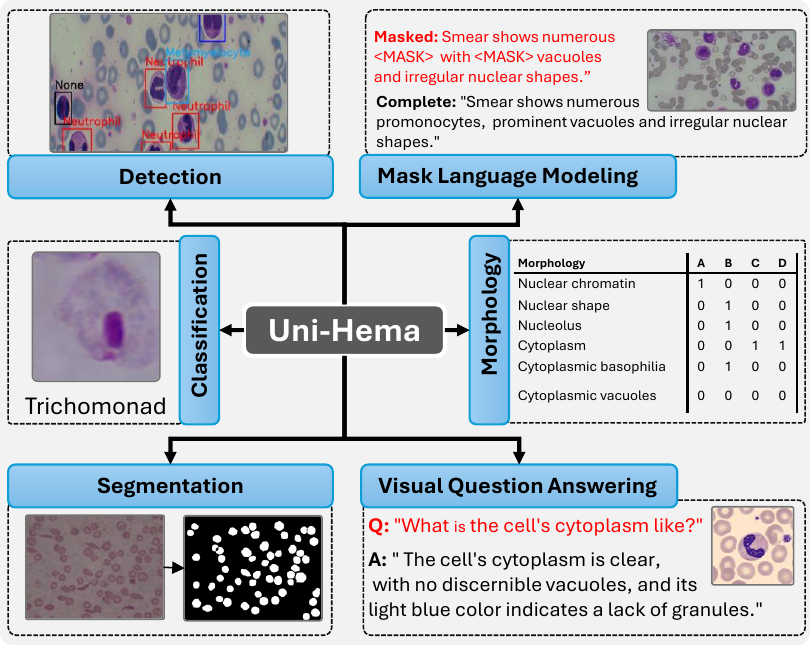}  
   \caption{\textbf{Uni-Hema}: A unified architecture supporting diverse digital hematopathology tasks; including cell detection, morphology prediction and  cell segmentation in both single-cell and complete field-of-view (FoV) images, as well as visual question answering, and masked language modeling.}

    \label{fig:yourlabel}
\end{figure}

Hematology, the study of blood and its disorders \cite{platt1969color}, relies heavily on peripheral blood film (PBF) examination for diagnosing conditions such as malaria, anemia, sickle cell disease, thalassemia, and leukemia \cite{bain2025blood}. 
Just to give an idea of scale of its need, malaria caused 263 million cases and 597,000 deaths in 2023, concentrating in the WHO African Region \cite{world2024malaria}, while sickle cell disease (SCD) affected 7.74 million people globally in 2021, resulting in  estimated 376,000 deaths \cite{GBD2021SickleCell2023} that year.
Leukemia further contributes to the global cancer burden with regional variations in incidence and mortality \cite{bray2024global}. 
These statistics underscore the critical need for accurate and scalable hematological diagnostics, as interpretation of blood films remains complex, prone to inter-observer variability \cite{tayebi2022automated}, and limited by shortages of trained hematologists and resources, specifically in low- and middle-income countries \cite{pain2024barriers,xing2023artificial}.

\begin{figure*}[h]
    \centering
\includegraphics[width=1\textwidth]{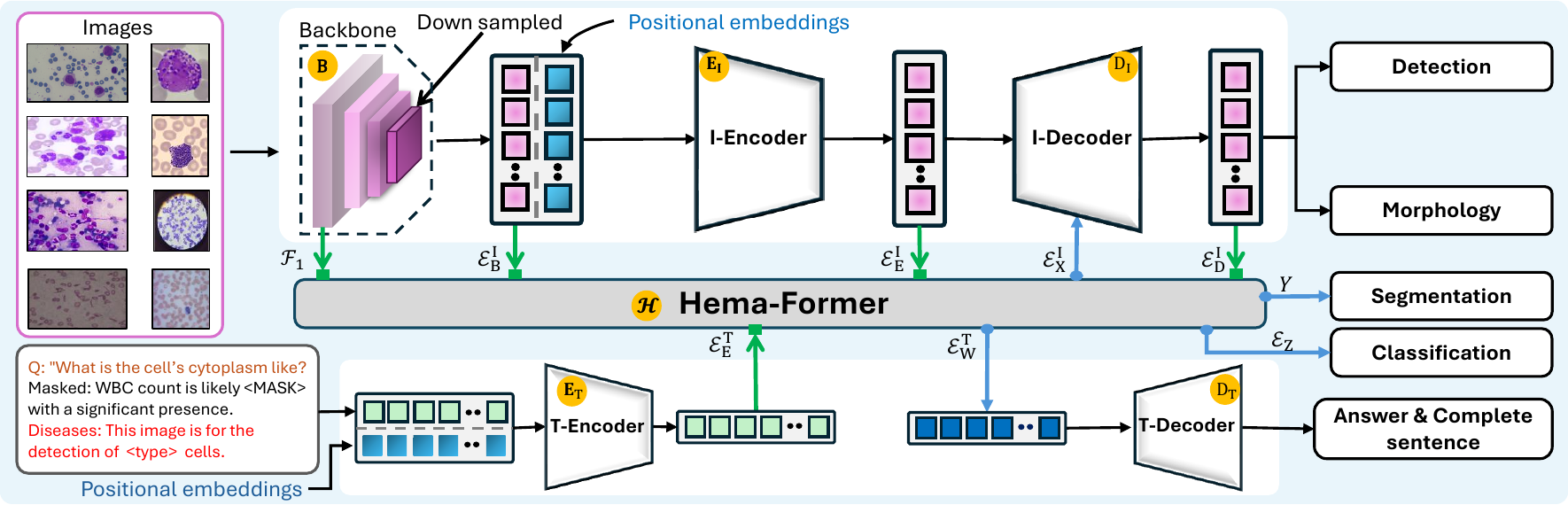}  
    \caption{Uni-Hema model architecture 
     comprises six principal modules: ($\mathbf{B}$) an image backbone for extracting spatial features, ($\mathbf{E}_\mathbf{I}$) an image encoder for hierarchical visual embeddings, ($\mathbf{D}_\mathbf{I}$) an image decoder for cell detection and morphology prediction; ($\mathbf{E}_\mathbf{T}$) a text encoder for extracting textual features with respect tasks,  ($\mathbf{D}_\mathbf{T}$) a text decoder for answer and sentence generation, ($\mathcal{H}$) and a Hema-former module, which serves as the core of the model by bridging visual and textual representations employing four submodules (see section~\ref{fig:hema_former}). 
    }

    \label{fig:uni_hema}
\end{figure*}

Recent advances in medical image analysis have transformed traditional blood based microscopy examination into the emerging field of digital hematopathology  \cite{hu2022artificial}. Previously,  most of the existing methods \cite{sultani2022towards, acevedo2020dataset, parasite-detection_dataset, kouzehkanan2021raabin, rehman2024large} were designed for single-task and single-disease applications, requiring separate datasets for each diagnostic purpose, thereby restricting scalability.
{Blood smear images further enhance this challenge due to overlapping cells, wide morphological diversity, and the need for multi-task processing, including detection, segmentation, classification, morphological interpretation, and visual–textual reasoning \cite{shahzad2024blood}.} 
Furthermore, such task-specific approaches may restrict clinical applicability, as they struggle to adapt to the diverse and complex demands of real-world healthcare environments.\cite{rajpurkar2023current, moor2023foundation}. 
To overcome this issue, Moor \textit{et.al}~\cite{moor2023foundation} have proposed Generalist Medical Artificial Intelligence (GMAI), which utilizes the recently developed medical foundation models \cite{wiggins2022opportunities} for the different medical tasks. Furthermore, the models in GMAI depend on natural language-based supervision \cite{moor2023med, tu2024towards,huang2023visual}.
Despite their success in  vision-language tasks, these foundation models are limited in addressing essential vision-centric problems like detection and segmentation \cite{chen2022unified, zhang2022glipv2}. 
The foundation models in digital pathology are primarily optimized for whole slide images (WSIs) of solid tissues captured at lower magnifications (5×, 40×) \cite{wang2024pathology, ochi2024pathology, wang2021transpath, chen2024towards}. In contrast, hematological diagnosis requires fine-grained, cell-level information at much higher magnifications (40×, 100×) \cite{acevedo2020dataset, alom2018nuclei, hehr2023morphological, rehman2025leveraging, koch2024dinobloom}. 
On the other hand some recently introduced hematology foundation models \cite{koch2024dinobloom, zedda2025reddino} are largely limited to single-cell tasks thus unable to handle realistic complexity of multiple cells (which might be overlapping or connecting), and even then are restricted by the cell type and tasks.

These limitations highlight a critical gap and emphasize the need for a \textit{unified} multi-task, multi-modal (image and text), multi-disease model tailored specifically to the demands of digital hematopathology (DHP).
Such model should unify visual and textual understanding for digital hematopathology by integrating preferably existing datasets and perform diverse tasks (classification, detection, segmentation, and morphological reasoning), enabling comprehensive, cell-level interpretation across multiple hematological conditions such as leukemia, malaria, and anemia.
{Developing unified models is challenging because it depends on having a complete benchmark for all the targeted tasks \cite{jin2024you}. 
Most publicly available hematopathology datasets are disease-specific and task-limited,  for example, focusing solely on malaria parasite detection \cite{sultani2022towards, parasite-detection_dataset, GBD2021SickleCell2023} or leukocyte classification \cite{acevedo2020dataset, kouzehkanan2021raabin, gan2024txl}, thereby lacking the diversity needed to generalize across multiple hematological disorders.}
 {In parallel, even in the pathology vision–language foundation models \cite{wang2022medclip, chen2024towards, ochi2024pathology, lu2023towards} rely on paired image and caption datasets, which are difficult to create in hematopathology because they require precise, cell-level, and clinically accurate descriptions \cite{tsutsui2023wbcatt,wang2021transpath}. The complexity increases further in field-of-view images of a microscope due to the wide variation in the number of cells, their types, and morphological patterns \cite{rehman2025leveraging}.}

To address these limitations, we propose \textit{Uni-Hema}, a unified model designed to for multi-task, multi-modal, and multi-disease of DHP.
Uni-Hema has been designed to perform detection, classification, segmentation, morphology prediction, and cell-level visual question answering by leveraging the existing datasets.  
To handle multiple tasks, image types, a unified model should learn and capture representations across multiple hierarchical levels, while effectively conditioning on features from the image and text input. 
Uni-Hema, embodies an image feature processing pipeline,  consisting of CNN  and transformer-based encoding and decoding, and text based encoder and decoder network. 
Information across these seemingly parallel networks is carried out by an information mixture module called \textit{Hema-Former}.
Once features are extracted, these are forwarded to different task heads, ranging from cell detection (localization) and cell classification, morphological feature prediction, binary image segmentation head, image classification, and answer \& sentence completion head. 
Our contributions are summarized below. 



\begin{itemize}
 \item We propose \textbf{Uni-Hema}, the first unified vision-language model for digital hematopathology that jointly performs multiple tasks such as detection, segmentation, classification, and morphology reasoning across the six diverse disorders and diseases, including leukemia, malaria, anemia, and sickle cell disease.
    
\item We have designed Hema-Former, a multi-modal feature fusion mechanism. It inputs feature representations of text and image at different hierarchical levels, and using attention mechanism generates features of different granularity appropriate for different tasks. These different feature generation strategies are implemented through learnable 4 sub-modules. (Section \ref{Hema-Former})

    
    \item We combine 46 diverse hematology datasets comprising 11 segmentation $(\approx221K)$, 18 detection $(\approx85K)$, and 17 classification $(\approx320K)$ datasets, along with curated 22K Question Answer pairs and 7K masked language modeling samples, establishing the most extensive multi-modal corpus for digital Hematopathalogy to date.
    
    
    \item The unified model allows to learn representations that benefit from detection, segmentation, and single-cell interpretation tasks. Therefore, even when trained once jointly for all the tasks, the results are better or comparable to the single-task, single-dataset SOTA methods. This trend is visible on unseen datasets as well. 
    
    
 \end{itemize}

\section{Related Work}
\label{sec:formatting}
The early approaches in digital hematopathology are limited to addressing single-task objectives, mostly focusing on single cell classification \cite{acevedo2020dataset, firat2024classification, chen2024sckansformer, matek2019single}, detection \cite{sultani2022towards, parasite-detection_dataset, tushabe2024image, quinn20186}, or segmentation \cite{depto2021automatic, koohbanani2020nuclick, shahzad2024anerbc, loddo2018mp, zheng2018fast, mohamed2012enhanced} independently. 
More recent research has extended toward dual-task and multi-task learning frameworks~\cite{rehman2024large, pal2024advancing, pang2025cellotype}, enabling the joint optimization of related tasks such as classification and segmentation to enhance diagnostic precision and computational efficiency. 
Driven by rapid advancements in deep learning, current research has increasingly focused on developing VLM's \cite{jia2021scaling, filiot2024phikon, lu2023towards}, foundation models~\cite{koch2024dinobloom, zedda2025reddino, ochi2024pathology, wang2022medclip}, and unified frameworks~\cite{li2023uni3dl, nehrdich2024one} capable of generalizing across tasks and modalities, moving beyond traditional task-specific designs.

\noindent\textbf{Vision Language Models (VLM's):}
The emergence of VLM's has bridged visual and textual modalities, enabling joint reasoning across computer vision and natural language processing tasks. Models such as CLIP~\cite{radford2021learning}, ALIGN~\cite{jia2021scaling}, and CoCa~\cite{yu2022coca} have pioneered task-agnostic pretraining by leveraging large-scale image–caption pair datasets to learn shared representation spaces, achieving strong generalization across diverse downstream tasks.
Despite these advances, the lack of large-scale, high-quality image–text datasets specific to the sub-domains of pathology continues to constrain their generalizability and downstream applicability.
Pathology oriented frameworks like CONCH \cite{lu2023towards}, Transpath \cite{wang2021transpath}, and Pathology foundation model (PFM) \cite{ochi2024pathology} introduce domain adaptation strategies to improve multi-modal alignment for tissue slides. However, their transferability to hematology remains limited due to hematology-specific data scarcity, cells morphological diversity, and the fine-grained cellular variations that different from pathology imaging modalities.


\noindent\textbf{Foundation model: } 
In the biomedical imaging domain, foundation models represent an effort to move beyond narrow task formulations by establishing transferable visual representations that can support a range of downstream analyses from tissue segmentation to cellular morphology classification. Approaches such as MedSAM \cite{MedSAM}, DinoBloom \cite{koch2024dinobloom}, and RedDino \cite{zedda2025reddino} demonstrate the growing potential of large-scale pretraining in medical and hematology contexts.
However, these models remain limited to single-cell analysis and textual integration,  lacking unified multimodal reasoning across detection and contextual reasoning.  While multimodal foundation models (e.g., UNI \cite{chen2024towards}, PFM \cite{ochi2024pathology}) emphasize image–text retrieval but lake in fine-grained biomedical understanding.

\noindent\textbf{Unified model:}
Recent advancements have accelerated the development of unified models capable of performing multiple tasks within a single framework. Models such as DINO-X~\cite{ren2024dino}, Uni-3DL~\cite{li2023uni3dl}, and Uni-Perceiver v2~\cite{li2023uni} exemplify this trend by incorporating multimodal inputs and modality-agnostic transformers to enable scalable, task-agnostic learning across diverse domains. However, in the medical imaging domain, unified architectures remain limited. Existing approaches, such as 
CelloType~\cite{pang2025cellotype} and  LeukemiaAttri \cite{rehman2024large} demonstrate progress toward multi-task learning by jointly performing dual tasks, but still rely on task-specific single dataset. Nonetheless, these methods remain constrained to specific diseases with annotated multitask datasets and lack the broader adaptability required for a unified model for multi-task, multi-disease, multimodal hematological analysis.
To address the above mentioned limitation of VLMs, foundation and unfied models for hematopathology, our proposed unified framework leverages diverse datasets to enable integrated learning against multiple tasks and hematology diseases.  

\begin{figure}[h]
    \centering
\includegraphics[width=0.95\linewidth]{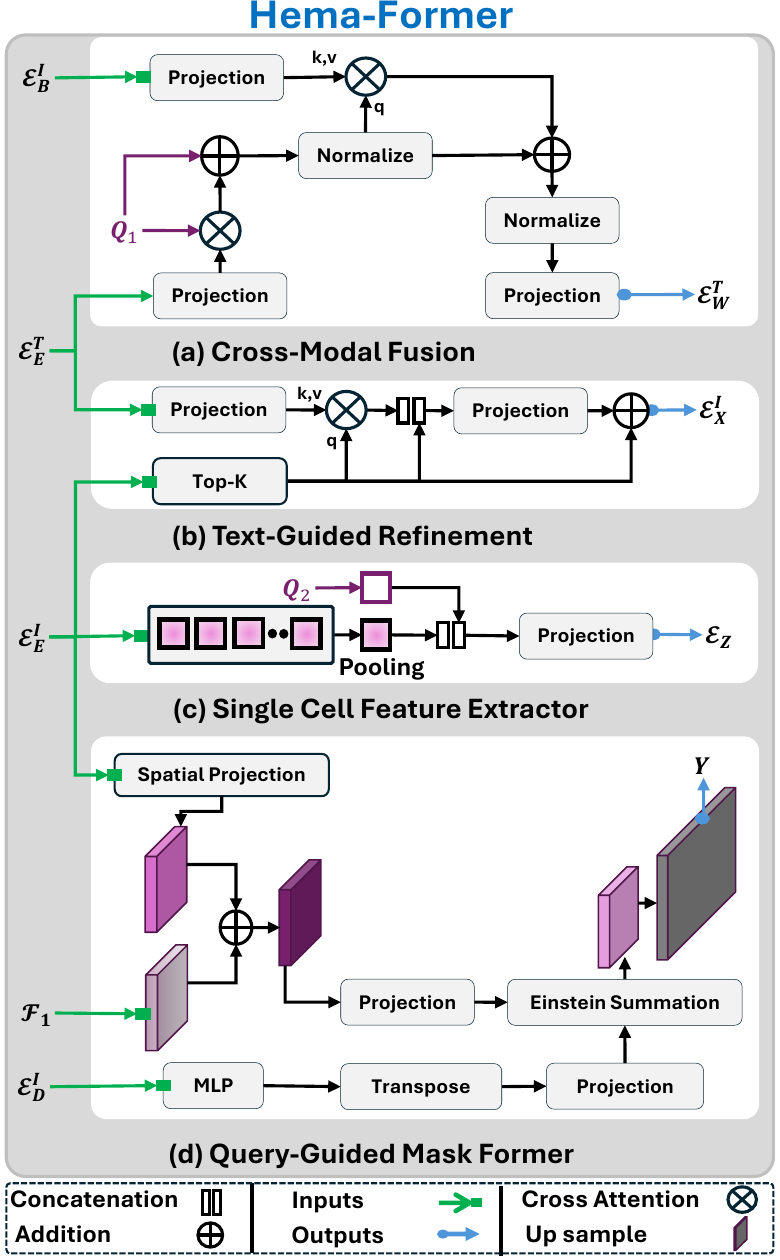}  
    \caption{
    Hema-Former sub-modules: (a) Cross modal fusion aligns text feature queries ($\mathcal{E}_\mathbf{E}^\mathbf{T}$) and image feature queries ($\mathcal{E}_\mathbf{E}^\mathbf{I}$) incorporate learnable queries ($\mathbf{Q}_1$) 
    (b) Text-guided refinement that updates  top-K object queries  
     through cross-attention with text queries  (c) Single cell feature extractor  aggregates mean-pooled encoder features ($\mathcal{E}_\mathbf{E}^\mathbf{I}$) with learnable query ($\mathbf{Q}_2$) for single-cell classification,  and (d) Query-guided mask former (QMF) module that utilizes the fused backbone features ($\mathcal{F}_1$) and spatially projected encoder features ($\mathcal{E}_\mathbf{E}^\mathbf{I}$) to predict segmentation masks ($\mathbf{Y}$),.
    }
    \label{fig:hema_former}
\end{figure}
\section {Methodology}
\subsection{Dataset Formation}

To train the unified model, we utilized 46 publicly available hematology datasets encompassing both single-cell and full-field microscopic images, covering a wide range of disease categories, including malaria, leukemia, anemia, sickle cell disease, and healthy samples. For segmentation or detection tasks, we generate a textual prompt for each image in the form: “This image is for the detection of  \textless type of disease \textgreater~of cells.”
For Question Answer, the prompt starts with Q: and for Masked language modeling, the prompt starts with mask: tokens. The single-cell classification relied solely on visual features. The details of datasets is given in Section  \ref{dataset}.

\subsection{Uni-Hema }
The architecture of Uni-Hema ($\mathcal{U}$) is illustrated in Figure \ref{fig:uni_hema}. It serves as an integrated vision–textual framework designed to address multiple tasks in digital hematopathology. The architecture comprises six principal modules: the backbone ($\mathbf{B}$), image encoder ($\mathbf{E}_{\mathbf{I}}$), text encoder ($\mathbf{E}_{\mathbf{T}}$), image decoder ($\mathbf{D}_{\mathbf{I}}$), text decoder ($\mathbf{D}_{\mathbf{T}}$), and hema-former module ($\mathcal{H}$)

We first outline the notation and basic feature extraction steps used in UNI-Hema. 
Given an input hematological image $\mathbf{I}$ and a text prompt $\mathbf{T}$, the image backbone $\mathbf{B}$ extracts   multi-scale spatial features $\{\mathcal{F}_i\}_{i=1}^{L}$, where each $\mathcal{F}_i \in \mathbb{R}^{c_i \times m_i \times n_i}$ represents the $i^{\text{th}}$ feature map with $c_i$ channels and spatial resolution $m_i \times n_i$ and `L' represents the number of levels.
These features are converted into a spatial embedding space $\mathcal{E}_{\mathbf{B}}^\mathbf{I} \in \mathbb{R}^{V_t \times M}$ and encoded by a pre-trained encoder $\mathbf{E_I}$ to produce contextualized visual representations $\mathcal{E}_{\mathbf{E}}^\mathbf{I} \in \mathbb{R}^{V_t \times M}$, where $V_t$ denotes the number of spatial tokens and $M$ represents the embedding dimension. In parallel, text encoder $\mathbf{E}_\mathbf{T}$ derives contextual embeddings  $\mathcal{E}_\mathbf{E}^\mathbf{T} \in \mathbb{R}^{L_t \times N}$ from the text prompt $\mathbf{T}$,  where $L_t$ denotes the number of textual tokens and $N$ is the text embedding dimension.
Finally, the image decoder $\mathbf{D_I}$ refines the $\mathcal{E}_{\mathbf{E}}^\mathbf{I}$  features into disease-informed object embeddings $\mathcal{E}_{\mathbf{D}}^\mathbf{I}$ for downstream reasoning.
\subsubsection{Image and Text Encoders:}

\noindent\textbf{Image Backbone:}
We employ a CNN-based ResNet  \cite{he2016deep}  backbone ($\mathbf{B}$) as the visual feature extractor. The $\mathbf{B}$ generates multi-level spatial feature  $\{\mathcal{F}_i\}_{i=1}^{L}$, capturing both fine-grained low-level information and high-level contextual details. These features are utilized for generating the segmentation masks, while projected spatial embeddings $\mathcal{E}_{\mathbf{B}}^{\mathbf{I}}$ serve as foundational representations for the other tasks.

\noindent\textbf{Image Encoder:}
To enhance contextual understanding and capture long-range dependencies between spatial embeddings, inspired by DINO~\cite{zhang2022dino}, a six-layer spatial image encoder $\mathbf{E_I}$ is integrated on top of backbone features. The $\mathbf{E_I}$ refines $\mathcal{E}_{\mathbf{B}}^{\mathbf{I}}$ into contextualized embeddings $\mathcal{E}{\mathbf{E}}^{\mathbf{I}}$ using self-attention, enabling the model to encode both global and local structural relationships. Global features support robust classification of single cells and morphology, while local features capture objectness in the microscopy field of view.

\noindent\textbf{Text Encoder:}
Textual branch of the~${\mathcal{U}}$ framework employs a transformer-based $\mathbf{E}_\mathbf{T}$ derived from T5~\cite{raffel2020exploring}. $\mathbf{E}_\mathbf{T}$ processes the text prompt $\mathbf{T}$ of hematological disease type, morphology-related questions, and masked sentences to produce contextualized textual embeddings $\mathcal{E}_{\mathbf{E}}^{\mathbf{T}}$. These embeddings capture underlying semantic relationships, enabling effective multimodal alignment and task-aware interaction with corresponding visual representations.
\subsubsection{Hema-Former:}\label{Hema-Former}
The proposed \textbf{Hema-Former} $({\mathcal{H}})$ integrates inputs from both text and image encoders to produce hierarchical, task-specific fused embeddings. For visual question answering and masked language modeling, it aligns textual and visual representations to enable multimodal reasoning. For detection and segmentation tasks, it focuses on token-level attention to enhance targeted object localization, while for pixel-level operations, it combines textual embeddings with fine-grained backbone features and image embeddings to generate accurate segmentation masks. Furthermore, a classification feature extractor enables text-independent image-level representation learning for classification.
Figure \ref{fig:hema_former} illustrates the architecture of the hema-former module, where the proposed hema-former bridges visual and textual representations through a set of integrated sub-modules. 

\noindent\textbf{ a) Cross-modal fusion (CMF):}
For effective alignment between textual and visual semantic embeddings, the proposed CMF computes cross-attention correlations between  $\mathcal{E}_\mathbf{E}^\mathbf{T}$, $\mathcal{E}_\mathbf{B}^\mathbf{I}$, facilitating unified understanding.  
Let the textual embeddings be $\mathcal{E}_\mathbf{E}^\mathbf{T} \in \mathbb{R}^{L_T \times N}$ and the visual features be $\mathcal{E}_\mathbf{B}^\mathbf{I} \in \mathbb{R}^{L_V \times M}$, learnable queries be about $\mathbf{Q}_2 \in \mathbb{R}^{L_V \times M}$. 
First, cross-attention is applied between $\mathbf{Q}_2$ and the projected textual embeddings ${\mathcal{E}_\mathbf{E}^\mathbf{T}}'' = \text{P}(\mathcal{E}_\mathbf{E}^\mathbf{T})$:
\begin{align}
\mathbf{J} &= \mathrm{Norm}(\mathbf{Q}_2 + \mathrm{CrossAttn}(\mathbf{Q}_2, {\mathcal{E}_\mathbf{E}^\mathbf{T}}'', {\mathcal{E}_\mathbf{E}^\mathbf{T}}''),
\end{align}
where $\mathrm{Norm}$ represents the normalization layer applied to stabilize features distribution.
Next, cross attention is applied between the  $\mathbf{J}$ and projected visual features $\text{P}(\mathcal{E}_\mathbf{B}^\mathbf{I})$ ,  then fused  with $\mathbf{J}$ and normalize as:
\begin{align}
\mathcal{E}_\mathbf{W}^\mathbf{T} &= \mathrm{Norm}(\mathbf{J} + \mathrm{CrossAttn}(\mathbf{J}, \text{P}(\mathcal{E}_\mathbf{B}^\mathbf{I}), \text{P}(\mathcal{E}_\mathbf{B}^\mathbf{I})),
\end{align}
where $\mathcal{E}_{\mathbf{W}}^{\mathbf{T}} \in \mathbb{R}^{L_t \times N}$ enriches the semantically aligned multimodal embeddings, enabling joint reasoning over visual and textual domains for the text decoder ($\mathbf{D}_T$) to generate language-driven tasks such as masked sentence completion and question answering.  



\noindent\textbf{ b) Text-guided visual refinement (TGVR): }
To enable the distinction between different types of hematological diseases images for the segmentation and detection tasks,  the Top-$K$ $(\mathbf{k})$ queries are separated from the  $\mathcal{E}_\mathbf{E}^\mathbf{I}$ based on the objectness, similar to  \cite{zhang2022dino}. 
For the disease-attend visual representation, the textual embeddings $\mathcal{E}_\mathbf{E}^\mathbf{T}$ are processed jointly with the $\mathbf{k}$ separated visual queries. 
To align their dimensions, a projection layer is first applied as ${\mathcal{E}_\mathbf{E}^\mathbf{T}}' = \text{P}({\mathcal{E}_\mathbf{E}^\mathbf{T}})$, 
where $\text{P}(\cdot)$ denotes the projection function. 
The resulting ${\mathcal{E}_\mathbf{E}^\mathbf{T}}'$ is then utilized as the key and value in a cross-attention operation, 
with  $\mathbf{k}$ serving as the query. 
The attended features obtained from this operation are concatenated with $\mathbf{k}$ and passed through an additional projection layer 
to generate the refined decoder queries  $\mathcal{E}_{\mathbf{X}}^{\mathbf{I}}$:
\begin{equation}
\mathcal{E}_{\mathbf{X}}^{\mathbf{I}} = \text{P}\left([\mathbf{k} \, || \, \text{CrossAttn}(\mathbf{k}, {\mathcal{E}_\mathbf{E}^\mathbf{T}}',{\mathcal{E}_\mathbf{E}^\mathbf{T}}')]\right),
\end{equation}
where $||$ indicates concatenation.
$\mathcal{E}_{\mathbf{X}}^{\mathbf{I}} \in \mathbb{R}^{D_t \times M}$ represent the disease guided top-$K$ queries for the image decoder ($\mathbf{D}_I$), to  producing disease-informed object embeddings $\mathcal{E}_{\mathbf{D}}^\mathbf{I}$ that enable detection and morphology estimation and assist segmentation. {$D_t$} represent he dimension of the Tok-$K$ queries,

\noindent\textbf{ c) Single cell feature extractor   (SCFE):}
 {To extract global image level feature (independent of the text input), we design a  SCFE module. In this module, a learned query is concatenated with the mean of the image embeddings and then passed through a projection layer to produce the image-level feature, denoted as $\mathcal{E}_{\mathbf{Z}} \in \mathbb{R}^{1 \times M}$. }

\noindent\textbf{ d) Query-guided mask former (QGMF):} 
This module generates the binary segmentation mask by integrating spatial features from the backbone and contextual embeddings from the image encoder. It further incorporates disease-informed object embeddings from the image decoder, following the design principles of Mask DINO~\cite{li2023mask}.

Specifically, for the input image $\mathbf{I}$, spatial features ${\mathcal{F}_1}$ extracted from backbone are fused with the image embedding $\mathcal{E}_{\mathbf{E}}^{\mathbf{I}}$, extracted from image-encoder. 
In our current implementation fusion is done by summation operation but could be replaced by any other operation. 
The fused features are then passed through a learnable projection layer to obtain $\mathbf{G_{\text{proj}}}$.
In parallel, the disease-informed object query embeddings $\mathcal{E}_\mathbf{D}^\mathbf{I}$ are processed through a multilayer perceptron (MLP), followed by transposition and projection via $\mathbf{P}$:
\begin{equation}
{\mathcal{E}_\mathbf{D}^\mathbf{I}}' = \mathbf{P} \left( \mathrm{MLP}(\mathcal{E}_\mathbf{D}^\mathbf{I})\right)^{\top}.
\end{equation}
These projected embeddings are then combined with the spatial feature maps $\mathbf{G}_{\text{proj}}$ through a cross-modal interaction implemented using Einstein summation ($\mathcal{S}$), resulting in segmentation logits:
\begin{equation}
\mathbf{Y} = \mathcal{S}({\mathcal{E}_\mathbf{D}^\mathbf{I}}', \mathbf{G}_{\text{proj}}),
\end{equation}
where $\mathbf{Y} \in \mathbb{R}^{\mathbf{D}_t \times m \times n}$.  For binary segmentation, the mask associated with the first embedding is selected as the final binary segmentation prediction $\mathbf{Y} \in \mathbb{R}^{1 \times m \times n}$.

\subsubsection{Image and Text Decoders:}
\noindent\textbf{Image decoder:}
The image decoder ($\mathbf{D}_{\mathbf{I}}$)  in Uni-Hema extends the DINO Detr \cite{zhang2022dino} architecture. 
The output $\mathcal{E}_{\mathbf{D}}^{\mathbf{I}}$ of the image decoder $\mathbf{D}_{\mathbf{I}}$  serves as a shared representation for the three tasks i.e., segmentation (through Query-Guided Mask Former) and supporting cell detection and morphology prediction tasks.

\noindent\textbf{Text decoder:}
The text decoder in Uni-Hema builds upon the transformer-based architecture~\cite{raffel2020exploring}. It generates textual outputs autoregressively, attending to both previously decoded tokens and the fused multimodal representations through cross-attention with the visual encoder features. This mechanism allows the model to perform domain-specific reasoning tasks such as masked sentence completion and single-cell-based question answering. 


\section{Experiments}

\subsection{Datasets}
\label{dataset}
We curated a large collection of publicly available hematology datasets, preserving existing annotations for consistency. In total, 18 datasets were used for detection, 11 for segmentation, and 17 for classification, totaling $\approx$ {0.7} million images across diverse diseases. 
Two morphologically annotated datasets \cite{rehman2024large, tsutsui2023wbcatt} are used to curate the visual Question Answering and masked language modeling tasks.


\noindent\textbf{Segmentation : }
The cell segmentation datasets encompass malaria-infected cell datasets such as Malaria Cell Images \cite{abbas2020detection}, ErythrocytesIDB \cite{gonzalez2014red}, and MP-IDB \cite{loddo2018mp}; white blood cell segmentation datasets including BBBC041Seg \cite{depto2021automatic}, NuClick \cite{koohbanani2020nuclick}, KRD-WBC \cite{ali4617448white}, WBC Image dataset \cite{zheng2018fast}, and the White Blood Cell dataset \cite{mohamed2012enhanced}; as well as anemia and red blood cell datasets   \cite{shahzad2024anerbc,elsafty20241}.

\noindent\textbf{Single Cell for classification: }
The single-cell (SC) classification datasets, comprising both white blood cells (WBCs) and red blood cells (RBCs), were collected from a diverse range of publicly available studies and repositories. These include peripheral blood smear datasets \cite{hehr2023morphological, matek2019single, acevedo2020dataset, kouzehkanan2021raabin, koohbanani2020nuclick, alipo2022dataset, ali4617448white, BCCD, bodzas2023high, Shenderov2019APL, pal2024advancing, mourya2019all} and bone marrow datasets \cite{matek2021highly}.
In total, it has approximately 350,000 high-resolution single-cell images were utilized, spanning nearly 45 morphologically distinct WBC and RBC classes.

\noindent\textbf{Cell Detection:}
Our collection includes several large-scale and diverse hematological microscopy full microscopic field of view datasets for the cell detection, such as the largest leukemia dataset \cite{rehman2024large} that includes morphology attributes, along with M5 \cite{sultani2022towards}, TXL-PBC \cite{gan2024txl}, BCCD \cite{BCCD}, Sickle-cell \cite{tushabe2024image}, Erythrocytes \cite{gonzalez2014red}, MP-IDB \cite{loddo2019mp}, PD-PF \cite{tek2009computer}, Plasmodium \cite{quinn20186}, Vivax \cite{hung2017applying}, Raabin \cite{kouzehkanan2021raabin}, AneRBC \cite{shahzad2024anerbc}, Bio-Net \cite{shams2024bio}, ThickBloodSmears \cite{yang2019deep}, and NIH-NLM-Thick PV \cite{kassim2021diagnosing}.
In total, this collection comprises approximately 85,000 images representing 30 distinct cell classes, covering a wide disease categories.

\noindent\textbf{MLM and VQA: }
To generate descriptive sentences from microscopy data, we employed medical language model,  BioMistral 7B~\cite{labrak2024biomistral} on  the WBCAtt dataset~\cite{tsutsui2023wbcatt} to  construct question–answer pairs to single-cell reasoning and multimodal understanding. Similarly, {Gemini 1.5} \cite{team2024gemini} was applied to full field-of-view images from the LeukemiaAttri dataset~\cite{rehman2024large},  leveraging morphological annotations to create both masked and fully descriptive captions. We named both datasets as WBCAtt-VQA and LeukemiaAttri-MLM, respectively. The generation process for both datasets is {semi-synthetic}, where detailed ground truth morphological annotations are utilized as prompts to guide descriptive synthesis. The generated QA and masked sentence are verified through context-based evaluation to ensure semantic coherence and linguistic accuracy. Note that this dataset is curated only for experimental purposes and is not recommended for medical use.  

\begin{table*}[h]
\caption{Performance of Uni-Hema across multiple tasks. Baseline methods are trained separately for each task and for each dataset. Uni-Hema, on the other hand, is trained only once and tested on the tasks without any re-training or fine-tuning. A “–” symbol indicates that a baseline model does not support the corresponding task. Values in bold and underlined denote the best and second-best results.}
\centering
\resizebox{\textwidth}{!}{%
\begin{tabular}{l|cc|ccc|ccc|c|c|c|c}
\hline
\textbf{Dataset} & \textbf{DINO} & \textbf{YOLO} &\textbf{U-Net} & \textbf{Nanonet}  & \textbf{TransNetR}  &
\textbf{DinoBloomS} &
\textbf{Dinov2S} &
\textbf{Dinov3s} & 
\textbf{Resnet}&\textbf{AttriDet}&\textbf{Ours}& \textit{Disease}\\
\hline

\rowcolor{lightgray}
\multicolumn{2}{l}{\textbf{\textcolor{Blue}{Cell Detection} }} & \multicolumn{11}{c}{\textbf{mAP$_{50}$}}\\
\hline

H\_40x\_C2 \cite{rehman2024large}   & 36.9 & \underline{37.3} &  - & - & - & - & - & - & - & 40.6 &\textbf{43.6} & Leukemia  \\
H\_100x\_C2 \cite{rehman2024large}  & 43.7 & \underline{44.2} & - & - & - & - & - & - & -& 44.4   &\textbf{49.8} & Leukemia \\
L\_40x\_C2 \cite{rehman2024large}   & \underline{36.6} & 34.9   & - & - & - & - & - & - & - & 35.4  &\textbf{40.7} & Leukemia \\
L\_100x\_C2 \cite{rehman2024large}  & \underline{38.2} & 38.1   & - & - & - & - & - & - & -& 36.2  &\textbf{45.6} & Leukemia \\
H\_1000x \cite{sultani2022towards}  & \underline{79.8} & 77.3  & - & - & - & - & - & - & - & -  &\textbf{83.1} & Malaria \\
L\_1000x \cite{sultani2022towards}  & \textbf{64.2} & 56.5  & - & - & - & - & - & - & - & - &\underline{62.4} & Malaria  \\
H\_400x \cite{sultani2022towards}   & \textbf{70.4} & 66.9  & - & - & - & - & - & - & - & - &\underline{69.0} & Malaria \\
L\_400x \cite{sultani2022towards}   & 58.3 & \textbf{59.9}  & - & - & - & - & - & - & - & - &\underline{54.5} & Malaria \\
Sickle Cell \cite{tushabe2024image}   & \textbf{73.6} & \underline{68.6}  & - & - & - & - & - & - & - & -  &67.0&  Sickle Cell \\
Parasites \cite{parasite-detection_dataset} & \underline{38.6} & \textbf{46.1} &   - & - & - & - & - & - & - & - &36.2 & Parasites  \\
BCCD \cite{BCCD}                      & \textbf{89.5} & \underline{88.1} &   - & - & - & - & - & - & - & - &87.8 & Normal \\
TXL \cite{gan2024txl}                 & \textbf{95.3} & \underline{94.9} &   - & - & - & - & - & - & - & -  &94.0& Normal \\
\hline

Mean & 60.4 & 59.4 & - & - & - & - & - & - & - & - & 39.1 & \textbf{61.1} \\
\hline

\rowcolor{lightgray}
\multicolumn{2}{l}{\textbf{\textcolor{Blue}{Single Cell Classification}}} & \multicolumn{11}{c}{\textbf{F1}} \\
\hline

Raabin \cite{kouzehkanan2021raabin}      & - & - & - & - & -  & \underline{98.0} & 93.7 & 93.8 &88.1&-&\textbf{98.8}&Normal\\
BMC\cite{matek2021highly}         & - & - & - & - & - & \underline{85.0} & 68.2 & 67.8 &64.7&-&\textbf{86.2}&Lymphoma\\
\hline

Mean & - & - & - & - & - & 91.5 & 81.0 & 90.8 & 76.4 & - & \textbf{92.5} & - \\
\hline

\rowcolor{lightgray}
\multicolumn{2}{l}{\textbf{\textcolor{Blue}{Segmentation}}} & \multicolumn{11}{c}{\textbf{Dice-Score}}\\
\hline

AneRBC-Anemic\cite{shahzad2024anerbc}  & - & - & 78.3  & 91.2 & \textbf{93.6} & - & - & - & - & - &\underline{93.4}&Anemia \\
AneRBC-Healthy\cite{shahzad2024anerbc} & - & - & 75.1  & 90.9 & \textbf{95.2} & - & - & - & - & - & \underline{94.1}& Normal \\
Elsafty \cite{elsafty20241}        & - & - & 93.4  & 98.3 & \underline{99.5} & - & - & - & - & - & \textbf{99.9}& Anemia \\
IDB2\cite{loddo2018mp}           & - & - & 91.5  & 33.1 & \textbf{92.1} & - & - & - & - & - & \underline{90.5}& Leukemia  \\
KRD\cite{ali4617448white}            & - & - & 93.3  & 86.7 & \textbf{94.9} & - & - & - & - & - & \underline{94.5}&  Unknown \\
MD-2019\cite{abbas2020detection}        & - & - & 75.2  & 84.7 & \textbf{86.7} & - & - & - & - & - & 77.6& Malaria  \\
\hline

Mean & - & - & 84.5 & 80.8 & \textbf{93.7} & - & - & - & - & - & \underline{91.7} & - \\
\hline

\rowcolor{lightgray}
\multicolumn{2}{l}{\textbf{\textcolor{Blue}{Cell Morphology (FoV)}}} & \multicolumn{11}{c}{\textbf{F1}}\\
\hline

H\_40x\_C2 \cite{rehman2024large} & - & - & - & - & - & - & - & - & - & 64.1 &\textbf{75.0}&Leukemia\\
H\_100x\_C2 \cite{rehman2024large} & - & - & - & - & - & - & - & - & - &71.1&\textbf{75.8}&Leukemia\\
L\_40x\_C2 \cite{rehman2024large} & - & - & - & - & - & - & - & - & - &53.2&\textbf{74.5}&Leukemia\\
L\_100x\_C2 \cite{rehman2024large} & - & - & - & - & - & - & - & - & - &61.8&\textbf{83.6}&Leukemia\\
\hline

Mean & - & - & - & - & - & - & - & - & - & 62.6 & \textbf{77.2} & - \\
\hline

\rowcolor{lightgray}
\multicolumn{2}{l}{\textbf{\textcolor{Blue}{Single Cell Visual QA}}} & \multicolumn{11}{c}{\textbf{BLEU-4}}\\
\hline
WBCAtt-VQA & - & - & - & - & - & - & - & - & - & - &\textbf{56.4}&Normal\\
\hline

\rowcolor{lightgray}
\multicolumn{2}{l}{\textbf{\textcolor{Blue}{Masked Language Modeling}}} & \multicolumn{11}{c}{\textbf{BLEU-4}}\\
\hline
LeukemiaAttri-MLM & - & - & - & - & - & - & - & - & - & - &\textbf{79.8}&Leukemia\\
\hline

\end{tabular}
\label{tab:main_table}
}
\end{table*}

\subsection{Implementation details}
In this work, we use {ResNet-50}~\cite{he2016deep} as the backbone, and adopt the {DINO-DETR}~\cite{zhang2022dino} architecture with six transformer layers in both the encoder and decoder, initialized with pretrained weights. 
For textual processing, we employ encoder-decoder based {T5}~\cite{raffel2020exploring} initialized with \texttt{t5\_base} weights. 
Training is performed in six sequential stages.
\noindent\textbf{Step 1:} Pre-train the image backbone and encoder on single-cell (SC) classification datasets for 24 epochs with a batch size of 32. 
\textbf{Step 2:} Pre-train the text encoder and decoder on semi-synthetic masked language modeling (MLM) and question-answering (QA) datasets for the medical context learning.  
\textbf{Step 3:} Jointly train the vision modules of the Uni-Hema on classification, segmentation, and detection datasets using two images per task per batch (total batch size = 6) for 12 epochs, updating the visual modules, including the TGR, SCFE, and QGMF modules, simultaneously. The rest of the modules ($\mathbf{E}_\mathbf{T}$ , $\mathbf{D}_\mathbf{T}$, CMF) are freeze.  
\textbf{Step 4:} Fine-tune the image decoder for 12 epochs on all detection and microscopy-field-of-view (FoV) morphology datasets.  
\textbf{Step 5:} Fine-tune the query-guided mask former (QGMF) on the complete segmentation dataset for 24 epochs.  
\textbf{Step 6: }Train the cross-modal fusion (CMF)  and $\mathbf{D}_\mathbf{T}$ on combined VQA and V-MLM datasets for 8 epochs,  all other modules and submodules freeze. 
The complete training process requires approximately 8 days on a single NVIDIA RTX 4090 GPU (the best available resource that we have). 
The details of the training configuration for each step are provided in the Supplementary material. 

\subsection{Results and Discussion}
For detection, we report comparisons with DINO~\cite{zhang2022dino} and YOLO~\cite{glenn_jocher_2022_7347926}. For segmentation, we include U-Net, NanoNet~\cite{jha2021nanonet}, and TransNetR~\cite{jha2024transnetr}. In single-cell classification, we compare against DINOv2S~\cite{oquab2023dinov2}, DINO-BloomS~\cite{koch2024dinobloom}, and DINOv3S~\cite{simeoni2025dinov3}. For multi-task comparison, the performance of AttriDet~\cite{rehman2024large} on FoV morphology and cell detection is reported on the LeukemiAttri dataset. To evaluate VQA and MLM task, in-house curated datasets WBCAtt-VQA and LeukemiaAttri-MLM are used. Other results and comparisons for detection (Faster R-CNN~\cite{ren2016faster}, Sparse R-CNN~\cite{sun2021sparse}, FCOS~\cite{tian2019fcos}), segmentation (U-Net, Attention U-Net~\cite{oktay2018attention}, and classification (TransPath~\cite{wang2021transpath}, CONCH~\cite{lu2023towards}, Phikon-v2~\cite{filiot2024phikon},  ResNet-50~\cite{he2016deep},  UNI~\cite{chen2024towards}) are all reported in the Supplementary Material.

Note that in the following experiments, each baseline method is trained on its corresponding dataset individually, whereas Uni-Hema is jointly trained on the entire training corpus.

\noindent\textbf{Detection Results:} We compare the performance of cell detection across multiple datasets with previous state-of-the-art fully supervised methods as shown in Table \ref{tab:main_table}. The results demonstrate that our proposed unified method achieves superior or comparable performance on cell detection tasks across various diseases, including leukemia, malaria, sickle-cell, and normal individuals. For sickle cell and parasite detection datasets, the number of training samples is very limited, making it difficult to learn dataset-specific patterns within a large corpus. In contrast, when training on individual datasets, the reduced complexity allows the model to learn more effectively.

\noindent\textbf{Single-Cell Classification Results:}  
{For a fair comparison, we use Uni-Hema as a feature extractor, obtaining $\mathcal{E}_{\textbf{X}}$ features from its SCFE module, and evaluate them against pretrained state-of-the-art (SOTA) feature extractors.}
These include non-medical-domain models~\cite{simeoni2025dinov3, oquab2023dinov2, he2016deep}, medical-domain models~\cite {wang2021transpath, filiot2024phikon, lu2023towards, chen2024towards}, and a hematology-specific vision foundation model designed for white blood cell images, DinoBloom~\cite{koch2024dinobloom}. 
\begin{figure*}[h]
    \centering
    \includegraphics[width=0.95\linewidth]{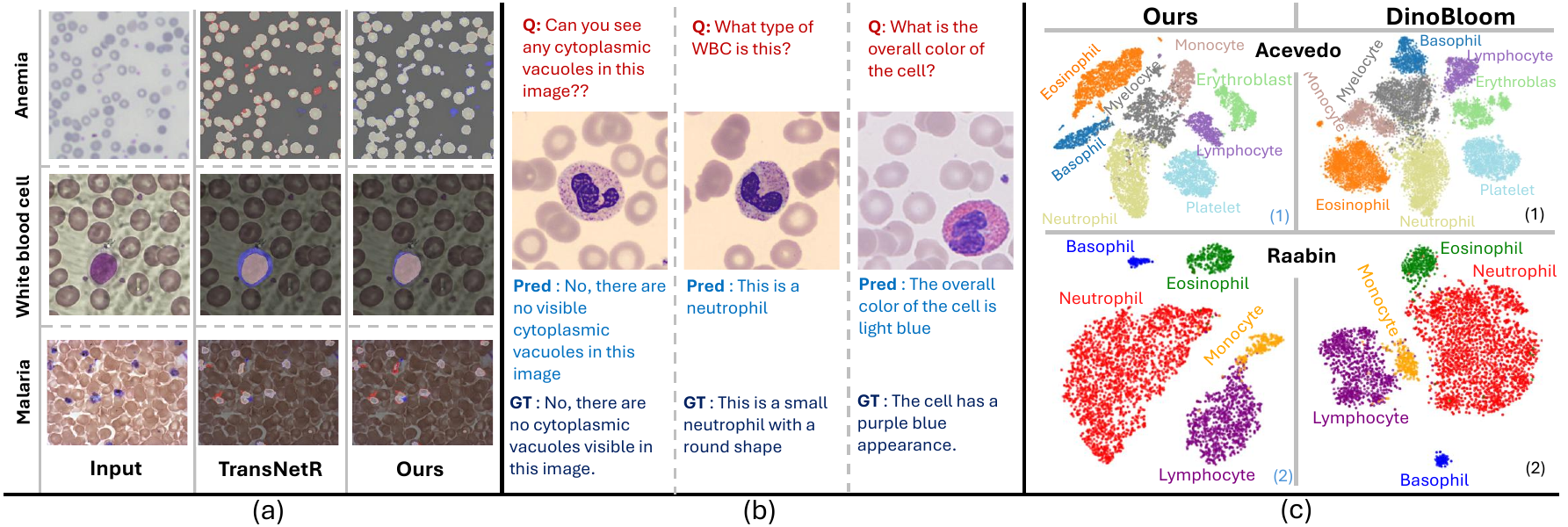}
    \caption{(a) Segmentation results of TransNetR and our method on anemia, malaria, and WBC images. TP, FP, and FN are shown in light yellow, blue, and red. Our method reduces false detections and improves localization, especially for anemia and WBC, while handling malaria robustly. (b) VQA outputs show contextually accurate predictions aligned with ground truth. (c) t-SNE plots of DinoBloom-S and Uni-Hema features on the Acevedo (8-class) and Raabin (5-class) datasets display clearer class separation with Uni-Hema.}
    \label{fig:segmentation}
    \end{figure*}
Experimental results Table~\ref{tab:main_table} demonstrate that Uni-Hema has superior performance compared to existing SOTA methods on both 21-class BMC ~\cite{matek2021highly}, and 5-class Rabin ~\cite{kouzehkanan2021raabin} dataset.


\noindent\textbf{Segmentation  Results:} To comprehensively assess segmentation performance, we compare Uni-Hema with leading fully supervised methods trained and tested on the particular dataset, including U-Net \cite{ronneberger2015u}, TransNetR~\cite{jha2024transnetr}, and NanoNet~\cite{jha2021nanonet}, as shown in Table~\ref{tab:main_table}. Our approach achieves competitive results, as reflected by the \textit{Dice score}. Notably, the reported performance corresponds to a single unified model evaluated across all datasets, without any task-specific retraining.

Instead of a SOTA decoder for upsampling, we rely on simple mask upscaling during training (step 3) and a small upsampler during fine-tuning (step 5) due to resource constraints. Although this may slightly affect edge sharpness, our results remain comparable to the state-of-the-art TransNetR method.
A qualitative comparison with TransNetR~\cite{jha2024transnetr} is illustrated in Figure~\ref{fig:segmentation}, while additional comparisons with other methods~\cite{ronneberger2015u, oktay2018attention, jha2021nanonet} are provided in the Supplementary Material.

\noindent\textbf {Morphology Results: } Cellular morphology across the microscopy field of view (FoV) is evaluated using the dataset from \cite{rehman2024large} which contains 6 morphological attributes.  As shown in Table~\ref{tab:main_table}, our method demonstrates superior performance compared to the corresponding SOTA multi-task, AttriDet~\cite{rehman2024large} method.
\begin{table}[h]
\centering
\caption{Comparison on unseen datasets: Uni-Hema demonstrates cross-domain adaptability across multiple tasks.}
\resizebox{\columnwidth}{!}{
\begin{tabular}{l|cccc|c}
\hline
\textbf{Dataset} & \textbf{DinoBloomS} & \textbf{Dinov2S} & \textbf{Dinov3S} & \textbf{ResNet} & \textbf{Ours} \\
\hline

\rowcolor{lightgray}
\multicolumn{6}{l}{\textbf{\textcolor{Blue}{Cell Classification (F1)}}} \\
\hline
Acevedo~\cite{acevedo2020dataset}     & \textbf{98.2} & 94.5 & 96.0 & 90.0 & \underline{98.1} \\
MiniSit~\cite{medmnistv2}             & \textbf{98.8} & 96.8 & 97.1 & 95.2 & \underline{98.6} \\
C-NMC~\cite{mourya2019all}            & 69.6 & \underline{71.0} & 68.0 & 64.9 & \textbf{72.8} \\
RV-PBS\_8~\cite{pal2024advancing}     & \underline{92.8} & 90.7 & 90.8 & 86.1 & \textbf{93.6} \\
\hline
Mean                                   & 89.9 & 88.2 & 88.0 & 84.1 & \textbf{90.8} \\
\hline

\rowcolor{lightgray}
\multicolumn{6}{l}{\textbf{\textcolor{Blue}{Detection (mAP\textsubscript{50})}}} \\
\hline
Bio-Net~\cite{shams2024bio}                               & - & - & - & - & 54.7 \\
Malaria~\cite{loddo2018mp, shams2024bio}                  & - & - & - & - & 78.5 \\
\hline

\rowcolor{lightgray}
\multicolumn{6}{l}{\textbf{\textcolor{Blue}{Cell Segmentation (Dice)}}} \\
\hline
BBBC041Seg~\cite{tsutsui2023wbcatt, depto2021automatic}   & - & - & - & - & 86.2 \\
\hline

\rowcolor{lightgray}
\multicolumn{6}{l}{\textbf{\textcolor{Blue}{Cell Morphology (Single) — F1-Macro}}} \\
\hline
WBCAtt~\cite{tsutsui2023wbcatt}                           & 87.6 & 87.8 & 91.4 & 90.2 & \textbf{91.6} \\
\hline

\end{tabular}
}
\label{tab:unseen_table}
\end{table}

\noindent\textbf{QA and MLM Results: } To broaden the capability of our proposed unified model, we incorporate two text-aligned visual understanding tasks, single cell visual Question Answering (VQA) and full field-of-view Visual Masked Language Modeling (VMLM), into the training framework.  As shown in Table~\ref{tab:main_table}, our {UNI-Hema} demonstrates strong performance across both tasks, achieving BLEU-4 scores of {56.4} for VQA, and {79.8} for V-MLM, respectively. These results highlight the model’s ability to bridge textual reasoning, including the visual interpretation in hematological contexts. Qualitative examples in Figure~\ref{fig:segmentation} further illustrate  the model’s predictions. 

\noindent\textbf{Results on unseen datasets: }
We evaluate Uni-Hema on unseen datasets, and the results are shown in Table~\ref{tab:unseen_table}. For the classification task, we follow DinoBloom~\cite{koch2024dinobloom} experimental setup.  Uni-Hema demonstrates overall superior performance withmean F1-score 90.8 . For cell detection task, it achieves a reasonable mAP@50 of 54.7 on the Bio-Net dataset~\cite{shams2024bio} and 78.5 on the Malaria dataset~\cite{loddo2018mp}. Furthermore, for segmentation on unseen datasets, Uni-Hema attains a Dice score of 86.2, highlighting its strong robustness across multiple tasks.  For single-cell morphological analysis, we follow the WBCAtt pipeline~\cite{tsutsui2023wbcatt}, treating the dataset as unseen for all feature extractor models \cite{oquab2023dinov2, simeoni2025dinov3, koch2024dinobloom, he2016deep} listed in Table~\ref{tab:unseen_table}. The results show that Uni-Hema achieves state-of-the-art performance on single-cell morphology prediction.

\noindent\textbf{Ablation studies: } In the single-cell classification task, we integrate features from both the image backbone and the single-cell feature composer. On the unseen domain-shift dataset of Acevedo \cite{acevedo2020dataset}, this integration improves accuracy from 97.9\% to 98.1\%, and on the C-NMC dataset \cite{mourya2019all}, from 68.8\% to 72.8\%. For the segmentation task, replacing simple interpolation-based upsampling with a small learnable upsampler network consistently enhances performance, improving segmentation results on the anemic dataset from 91.8\% to 93.4\% and on the healthy dataset from 92.8\% to 94.1\%, with similar gains observed across other datasets. Additional results are given in the supplementary material.
\section{Conclusion}
In this work, we present Uni-Hema, a unified model for multi-task, multi-disease, and multi-modal digital hematopathology. Uni-Hema integrates detection, segmentation, classification, morphological reasoning, and visual–textual understanding in single architecture. A dedicated Hema-Former module aligns hierarchical visual features with task-driven textual features for contextual understanding. We further curate VQA and MLM datasets specialized for hematopathology to support the capabilities. Trained on 46 heterogeneous datasets with a progressive strategy, Uni-Hema matches or surpasses state-of-the-art single-task models. 
{
    \small
    \bibliographystyle{ieeenat_fullname}
    \bibliography{main}
}
\end{document}